\documentclass[letterpaper, 10 pt, conference]{ieeeconf}
\IEEEoverridecommandlockouts

\usepackage{color}
\usepackage{subfiles}
\usepackage{graphicx}
\usepackage{float}
\usepackage{placeins}
\graphicspath{{Figures/}}

\begin{document}

\title{\LARGE \bf
Design and Experiments with a Robot-Driven Underwater Holographic Microscope for Low-Cost \emph{In Situ} Particle Measurements 
}

\author{Kevin Mallery$^{1*}$, Dario Canelon$^{1*}$, Jiarong Hong$^{1*}$, and Nikolaos Papanikolopoulos$^{2*}$% <-this % stops a space
\thanks{*This work was supported by National Science Foundation through grant \#IIS-1427014.}% <-this % stops a space
\thanks{$^{1}$ Department of Mechanical Engineering,
        University of Minnesota.
        }%
\thanks{$^{2}$ Department of Computer Science \& Engineering,
        University of Minnesota.
        }%
}

\maketitle

\begin{abstract}
%\boldmath
%Microscopic analysis of micro particles \emph{in situ} in diverse water environments is necessary for monitoring water quality and localizing contamination sources. Conventional sensors such as optical microscopes and fluorometers are restricted to small sample volumes and are unable to simultaneously capture all pertinent details of a sample such as particle size, shape, concentration, and three dimensional motion. In this article we propose a novel and cost-effective robotic system for mobile microscopic analysis of particles \emph{in situ} at various depths which are fully controlled by the robot system itself. A miniature underwater digital in-line holographic microscope (DIHM) performs high resolution measurements necessary to image individual organisms while movement allows measurement of particle distributions at a large scale of the body of water. The main contribution of this work is the creation of a novel, low-cost, comprehensive, and small underwater robotic holographic microscope that can assist in a variety of tasks in environmental monitoring and overall assessment of water quality. The resulting system provides some unique capabilities such as expanded and systematic coverage of large bodies of water at a low cost. Several challenges such as the stability of the sensor as the robot moves through the body of water are addressed to satisfy the aforementioned goals.

Microscopic analysis of micro particles \emph{in situ} in diverse water environments is necessary for monitoring water quality and localizing contamination sources. Conventional sensors such as optical microscopes and fluorometers often require complex sample preparation, are restricted to small sample volumes, and are unable to simultaneously capture all pertinent details of a sample such as particle size, shape, concentration, and three dimensional motion. In this article we propose a novel and cost-effective robotic system for mobile microscopic analysis of particles \emph{in situ} at various depths which are fully controlled by the robot system itself. A miniature underwater digital in-line holographic microscope (DIHM) performs high resolution imaging of microparticles (e.g., algae cells, plastic debris, sediments) while movement allows measurement of particle distributions covering a large area of water. The main contribution of this work is the creation of a low-cost, comprehensive, and small underwater robotic holographic microscope that can assist in a variety of tasks in environmental monitoring and overall assessment of water quality such as contaminant detection and localization. The resulting system provides some unique capabilities such as expanded and systematic coverage of large bodies of water at a low cost. Several challenges such as the trade-off between image quality and cost are addressed to satisfy the aforementioned goals.

%across a large-scale body of water

%This system allows high resolution measurements, necessary for smaller scale features, and movement to gather data over larger geographical spaces. This combination of 

%Microscopic analysis of micro particles in \emph{in situ} environments is necessary for monitoring water quality and localizing contamination sources. Conventional sensors such as optical microscopes and fluorometers are restricted to small sample volumes and are unable to simultaneously capture all pertinent details of a sample such as particle size, shape, concentration, and three dimensional motion. We propose a miniature underwater digital in-line holographic microscope (DIHM) mounted on a robotic platform as a cost-effective system capable of performing data collection \emph{in situ}.
%in order to facilitate necessary data in the field.
\end{abstract}


%% Field Robots, Distributed Robot Systems, Robotics in Agriculture and Forestry.
%\begin{IEEEkeywords}
%IEEEtran, journal, \LaTeX, paper, template.
%\end{IEEEkeywords}

% \fi
\IEEEpeerreviewmaketitle

\section{Introduction}
%\blindtext

\begin{figure}[!ht]
\centering
\includegraphics[width=1\columnwidth]{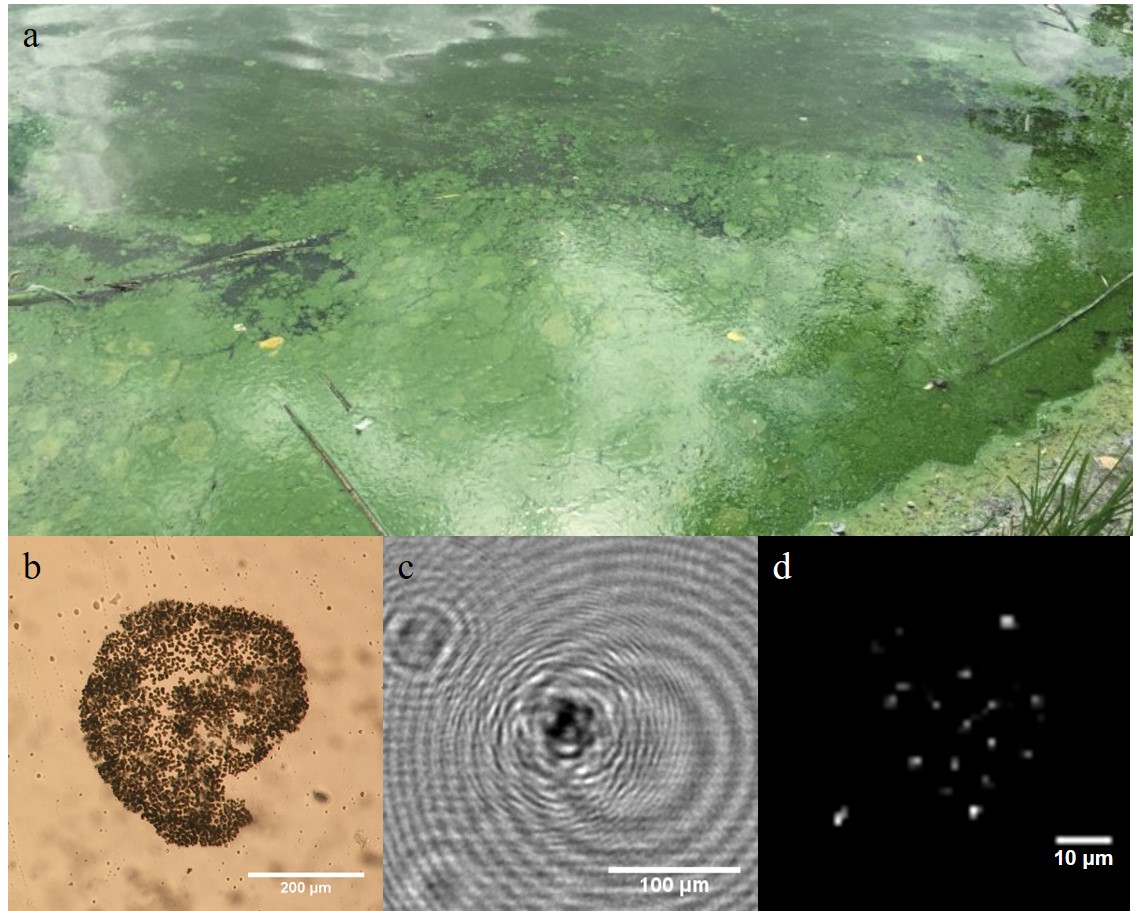}
\caption{(a) Macroscopic image of HAB in Powderhorn Lake, Minneapolis, MN. (b) \emph{M. aeruginosa} colony sampled from (a) and imaged using 400x optical microscope. (c) Recorded hologram of \emph{M. aeruginosa} colony using a laboratory DIHM. (d) In-focus image of (c) processed using the methods of this paper showing that the colony is comprised of individual cells. Photo credit: Jiaqi You.}
\label{fig:bloom_local}
\end{figure}

Water quality of aquatic environments is of critical importance for a variety of economic, environmental, and public health reasons. The economies of many regions around the globe are closely linked to the ability to use water resources for fishing, crop irrigation, and tourism. There are several sources of contamination that can cause millions of dollars in damages including harmful algal blooms (HABs) and oil spills \cite{U.S.EnvironmentalProtectionAgency2015}.
% Note: EPA is only for algae
The scale of HABs and oil spills can range from meters to kilometers and yet are made up of discrete microscopic objects. HABs often consist of colonial algae such as \emph{Microcystis aeruginosa} (Fig. \ref{fig:bloom_local}) which has a colony size on the order of 100 \(\mu\)m \cite{Rowe2016}. Oil droplets broken up and diffused by turbulence may have a similar scale \cite{Murphy2015}.
% Need more current method discussion here

Current methods for measuring such phenomena -- including fluorometric cytometry and satellite imaging -- are unable to measure across a wide range of length scales, often cannot differentiate between similar classes of particles, or cannot image beneath the surface. Because of this, physical sampling is still the gold standard measurement technique but is hindered by the need to process the samples in a laboratory and the difficulty of acquiring the samples. Digital in-line holographic microscopy is a 3D imaging technique which is becoming increasingly popular for underwater microscopy due to its broad applicability. However, high costs for commercial systems (approximately 40,000 USD) in addition to size and weight limitations provide a high barrier preventing more researchers from utilizing this technology.  

\begin{figure}[!t]
\centering
\includegraphics[width=1\columnwidth]{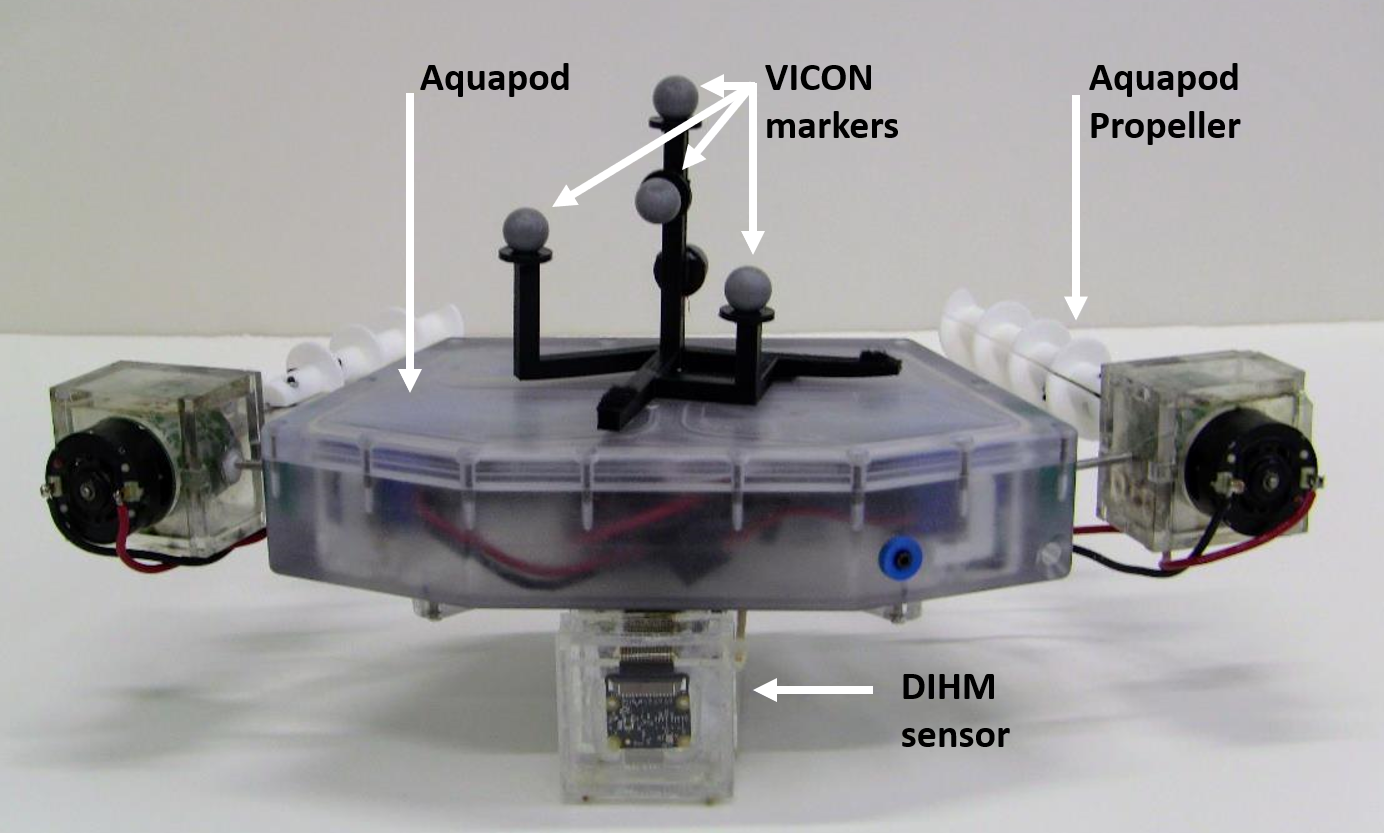}
\caption{DIHM mounted on the Aquapod prior to deployment.}
\label{fig:aquapod_dihm}
\end{figure}

In the following sections we present a novel system that allows the \emph{in situ} measurement of various environmental parameters of interest to the study of microorganisms and HABs. The system proposed is able to carry out measurements at both the small scales required to measure individual organisms (i.e., 1 \(\mu\)m) and also at the scales of the much larger bloom (i.e., 1 km). The system (Figure \ref{fig:aquapod_dihm}) consists of a digital in-line holographic microscope (DIHM) integrated into an autonomous amphibious vehicle -- the Aquapod -- in order to make high resolution measurements including particle size, concentration, identification, and three-dimensional motion as the Aquapod moves throughout a large environment of interest. We discuss the design of the DIHM in Section \ref{dihm}, placing a particular emphasis on decreasing the cost of the system and also addressing challenges such as the vibrations to the sensors as the robot propels itself. Section \ref{aquapod} covers the Aquapod robotic platform as it pertains to this work and how several platform limitations impact the overall system design. In Section \ref{experiments} we describe two experiments that were carried out as laboratory validation of our methods and in-field testing of our system and we discuss the results of the experiments. Finally, in Section \ref{future_work} we describe possible avenues for future research using this system.

\section{Background and Related Work} \label{background}

Digital in-line holographic microscopy (DIHM) is an optical imaging method based on the holographic method first described by Gabor in 1948 \cite{Gabor1948} to digitally reconstruct a 3D image of microscopic objects from a 2D recording. Of the many variants of holographic imaging, digital in-line holography is the simplest, only requiring a laser and a camera. The laser produces a spatially and temporally coherent light source in the form of a collimated beam. As the light propagates through a medium, particles in the field of view rescatter the incident light, producing a spherical wave. The interference of the spherical wave with the uncontaminated collimated wave is recorded by the camera, encoding the phase information of the spherical wave. After recording, the image can be digitally refocused (reconstructed) through convolution with the  Rayleigh-Sommerfeld diffraction kernel. These reconstructions can be used for species identification \cite{ElMallahi2013}, sizing \cite{Gopalan2008}, and particle tracking \cite{Katz2010}. The biggest advantage of holography over other imaging techniques such as traditional microscopy is that standard microscopes are limited to a narrow depth of focus while holography is capable of refocusing over a much wider volume (dof of 1 cm vs 1 \(\mu\)m). This substantially reduces the impact of the sensor on the fluid being measured and allows for a larger sampling volume, ensuring representative samples even under low concentration conditions.

The quality of the results that can be extracted from a recorded hologram is strongly dependent on a number of features of the imaging system. These include the presence of scratches or dust in the optical path, vibrations of the sensor, the beam profile of the illuminating laser, and the use of a microscopic objective or other lenses to adjust the region of interest. The spatial resolution of the measurements is dependent on the pixel size of the camera (usually 1-5 \(\mu\)m) and will be equal to that size in the special case when no objective lens is used (lensless holography). The temporal resolution is determined by the camera frame rate and shutter speed. In order to prevent motion blurring, the shutter speed must be less than the time taken for an object to pass through one pixel. DIHM data becomes increasingly noisy as the concentration increases. The non-dimensional shadow density (\(s_d=n_xLd^2\)) generally should not exceed 0.1 where \(n_x\) is the number density, \(L\) is the sample thickness, and \(d\) is the particle diameter \cite{Malek2004}. %

Other scattering based particle sensors such as many turbidity sensors and fluorometric cytometers utilize the spectra of light scattered from a single point and thus lack the spatial information inherent to an imaging method such as DIHM. The reduced sample volume necessary for these methods often requires some microfluidic sampling which may bias the sample composition, size distribution, or behavior in a way that ideal \emph{in situ} methods should not. Furthermore, these methods require careful calibration to account for the variations in scattering properties of different samples.
Optical microscopes are well suited for particle identification and counting as a trained operator (or algorithm) can perform these tasks directly from the recorded images without any need for calibration. Furthermore, high speed cameras enable monitoring of the motility of microorganisms. However, the limited depth of focus of conventional microscopes means that they are limited to 2D measurements limiting the measurement volume. In cases of rarefied samples, a very large sample volume may be necessary to detect the objects of interest. Such situations may occur in the deep ocean or in extraterrestrial environments. 3D microscopy techniques such as scanning confocal microscopy require a complex mechanical scanning mechanism and often require sample preparation via chemical staining. In addition to the mechanical complexity, scanning methods are unsuitable for measuring dynamic samples that change faster than the scanning speed. By comparison, DIHM requires no moving parts, can image at high speeds, and is non-destructive and non-invasive making it a very powerful measurement technique for \emph{in situ} studies.

There have been numerous previous applications of holography for underwater measurements. Talapatra et. al. \cite{Talapatra2012} used a submersible holographic device to measure the size and spatial distribution of particles in a 15 m water column. They also simultaneously measured the mean shear strain and turbulence dissipation rate by tracking the imaged particles. Lindensmith et. al. \cite{Lindensmith2016} used a  holographic microscope to detect life in frozen sea ice with proposed applications in extraterrestrial exploration. Deep sea holographic microscopes are sold commercially by 4Deep Inwater Imaging and Sequoia Scientific. Both systems produce high quality holograms and are capable of particle sizing, counting, and morphological species recognition at depths of over 100 m. Despite their utility, the large size (over 70 cm long), weight (7-27 kg), and cost (40,000 USD) of these systems limit potential applications.

In addition to the \emph{in situ} applications, laboratory studies have shown the power of DIHM as a measurement technique. High speed DIHM can be used for particle tracking velocimetry (PTV) \cite{Katz2010} which has applications in both flow measurement and behavior analysis. Holography has been used to examine the changes in the complex swimming behaviors of dinoflagellates in response to the presence of prey \cite{Sheng2007}. Other behavioral studies have examined the near-wall swimming behavior of bacteria \cite{Molaei2014} to relate swimming patterns to the local fluid vorticity \cite{Chengala2010}. DIHM has also been applied to automated detection and classification of organisms. El Mallahi et. al. \cite{ElMallahi2013} trained a support vector machine (SVM) classifier to automatically identify \emph{Giardia labmlia}, a waterborne parasite.

Despite the aforementioned advantages of DIHM, a fundamental drawback shared with most other techniques is that it operates on scales of less than 1 cm, while the properties of interest (e.g.,  extent of HAB) may vary at the scale of meters or kilometers. High resolution measurements are necessary for  efficient and effective monitoring of biological or chemical data but must be paired with a method for traversing the large three-dimensional environment in order to best understand the evolution of large scale features such as algal blooms \cite{Anderson2012}. This is one of the major innovations of this work since the redesigned Aquapod-based system  helps us achieve these mobility and coverage goals. 

%Fluid samplers are often used for studies of lakes and rivers, such as in the work of Villar-Argaiz et al. \cite{Villar-Argaiz2001} where monitoring the amount of phosphorous and nitrogen compounds was needed to determine factors that affected the growth of phytoplankton in lakes.  This type of sampling proves to be very resource intensive as the samples must be gathered under human supervision. Furthermore, there can be additional costs for transportation, preparation, and deployment -- all of which contribute to the initial costs associated with data gathering and can contribute greatly to the quantity and quality of the data that is produced. Cases like these would benefit greatly from the automation of data collection, as is available through autonomous robotic platforms, such as the Aquapod \cite{Carlson2011b} (Figure \ref{fig:aquapod-callout}).
%\cite{Dhull2012}

% ESP and Neumann
Employing robotic platforms to move sensors into desirable measurement locations is frequently being looked at as a way to gather data in a wider range of scenarios, better understand and gain insight on the environment through distributed and scaleable deployments, and augment human capabilities by leveraging automation in tasks that were not previously feasible. In events from the Horizon Oil Spill to the Fukushima Nuclear Disaster, robotic platforms have permitted environmental monitoring and the completion of vital tasks while reducing the need to place human resources in danger. In the case of environmental monitoring, the long-term objective is to deploy a number of robots to carry out sensory tasks in an automated fashion, sometimes spanning days or weeks, with occasional human interaction for maintenance or supervision.

In some cases, researchers have turned to robotic systems to deploy highly specialized and specific sensors to strategic locations to optimally gather data of the underlying physicochemical properties of the environment \cite{Neumann2012}. Utilizing small mobile platforms that integrate proportionately smaller sensors allows the active collection of information in a way that is not available by laboratory testing of samples or by stationary sensors. Furthermore, active sensors that process data \emph{in situ} permit measurement and detection, enabling a feedback loop based off live data to more intelligently deploy resources and gather the needed data in a more efficient and accurate manner \cite{Neumann2012}. 

Greenfield et al. have developed a robotic platform to perform laboratory-type measurements of algae in the field \cite{Greenfield2008}. This special machine, called the Environmental Sampling Processor (ESP), is a marine surface vessel which concentrates and filters physical samples to then utilize DNA probing in order to perform tasks such as species identification, particulate concentration, among others.

To our knowledge, there has been no previous integration of a DIHM with a robotic system. This represents a major opportunity to leverage the powerful 3D microscopic imaging capabilities of DIHM with the mobility and autonomy of a robotic platform. Compared to prior uses of DIHM, a miniature and cost-effective sensor is particularly valuable as it would enable a fleet of small distributed robots to actively explore the environment in stark contrast with the large individual ships currently used for underwater DIHM.

\section{Digital In-Line Holographic Microscope} \label{dihm}

\begin{figure}[!ht]
\centering
\includegraphics[width=1\columnwidth]{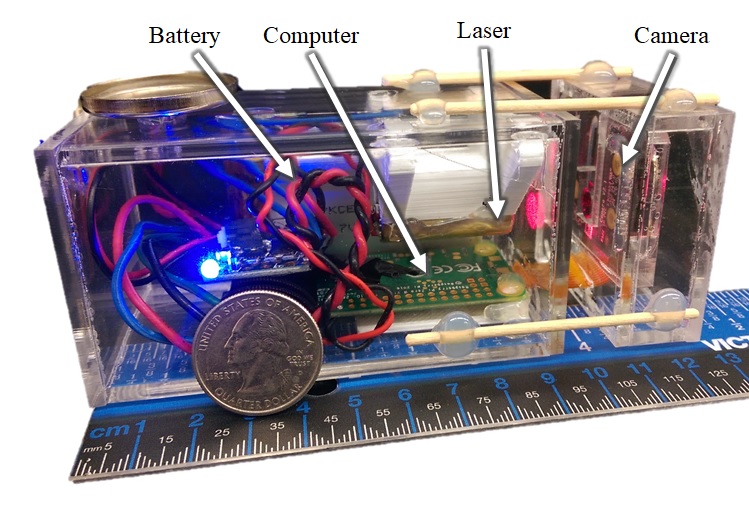}
\caption{DIHM system with labeled components.}
\label{fig:labeled_dihm}
\end{figure}

%\FloatBarrier

%\blindtext
%The DIHM then uses an optical imaging method based on the holographic method first described by Gabor in 1948 \cite{Gabor1948} to perform relevant data gathering. The method relies on using recorded 2D light interference patterns to digitally reconstruct the three dimensional light field. Of the many variants of holographic imaging, digital in-line holography is the simplest, only requiring a laser and a camera.

%The laser produces a spatially and temporally coherent light source in the form of a collimated beam. As the light propagates through a medium, particles in the field of view will rescatter the incident light, producing a spherical wave. The interference of the spherical wave with the uncontaminated collimated wave is recorded by the camera, encoding the phase information of the spherical wave. Convolution with the  Rayleigh-Sommerfeld diffraction kernel enables digital refocusing of the recorded image. These reconstructions can be used for species identification \cite{ElMallahi2013}, sizing \cite{Gopalan2008}, and particle tracking \cite{Katz2010}. The advantage of holography over other imaging techniques such as traditional microscopy is that standard cameras are limited to a narrow depth of focus while holography is capable of refocusing over a much wider volume (dof of 1 cm vs 1 \(\mu\)m). This substantially reduces the impact of the sensor on the fluid being measured and allows for a larger sampling volume, ensuring representative samples even under low concentration conditions.

In contrast with other DIHM sensors which emphasize image quality, 
our sensor design emphasizes cost, size, and simplicity while recording holograms of sufficient quality for several pertinent applications. 
We rely on our processing approach to compensate for the higher noise resulting from the use of cheaper and smaller components.
Figure \ref{fig:labeled_dihm} shows the complete sensor. The primary components are the laser, camera, acquisition computer, and battery.

The laser is a Quarton VLM650-11-LPA laser diode with a 650 nm wavelength and 3 mW power. The camera is a Raspberry Pi Camera Module NoIR V2 which has a maximum resolution of \(3296 \times 2512\) pixels, and a pixel size of 1.12 \(\mu\)m. Images are captured and stored using a Raspberry Pi Zero W single-board computer which also powers the camera and laser. All components are enclosed in an acrylic enclosure which is waterproof at depths up to 5 m. In order to improve size and reliability, the enclosure is permanently sealed and cannot be accessed by a user. Charging and data transfer are performed using an Qi inductive charging pad and 802.11 wireless connection.

The frame rate and field of view are interconnected and limited by bandwidth considerations when saving images. Increasing in the field of view requires decreasing in the image capture rate. For the experiments in this paper, an image size of \(2.3 \times 2.3\) mm (\(2048 \times 2048\) pixels) was recorded at 1 frame per second. The shutter speed was 1 \(\mu\)s. The depth of the sample volume (distance between imaging windows) is 10 mm, corresponding to a sample volume of 53 \(\mu\)L.

%\begin{figure}[!t]
%\centering
%\includegraphics[width=1\columnwidth]{Figures/DIHM_components.jpg}
%\caption{DIHM system with labeled components.}
%\label{fig:labeled_dihm}
%\end{figure}

The total cost of the system is \$200 (excluding labor) which is an order of magnitude reduction from previous systems and two orders less than commercial systems. The total size of this sensor is \(5 \times 5 \times 13\) cm, with the majority needed for the computer and battery. The sensor can be operated remotely via a broadcast 802.11 network signal and has a battery life of 3 hours. For underwater operation, recording is initiated via the 802.11 network prior to submerging the sensor. Image recording proceeds independent of the network connection while submerged until the trial is complete and the network connection is reestablished.

%\textcolor{red}{Need section on processing method}

\begin{figure}[!h]
\centering
\includegraphics[width=1\columnwidth]{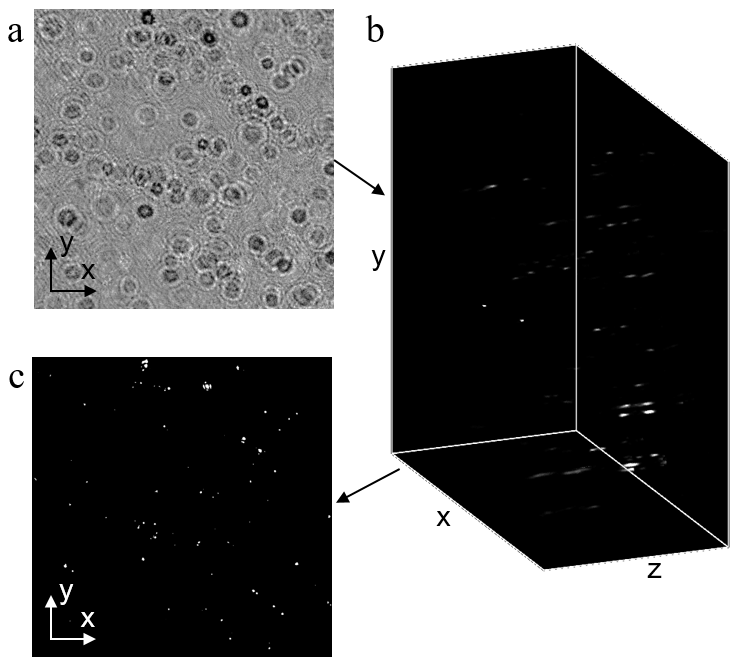}
\caption{Hologram processing steps. (a) Enhanced hologram of microbubbles. (b) 3D reconstruction. (c) Binarized 2D projection.}
\label{fig:holo_processing_steps}
\end{figure}

%The hologram processing is inspired by the compressive holographic approach of Endo et al. \cite{Endo2016} who use total variation (TV) regularization to solve an inverse problem using an iterative  algorithm to solve for the 3D object that best produces the recorded hologram. Our approach uses the fused lasso regularizer which enforces a sparse and smooth result. 
The holograms are processed using the RIHVR approach \cite{Mallery2019} which solves an inverse problem using an iterative algorithm to determine the 3D object that best produces the recorded hologram. Fused lasso regularization is used to ensure the result is both sparse and smooth.
The GPU-accelerated algorithm takes 21 seconds to process one \(2048\times2048\) pixel hologram with 40 reconstruction planes. Prior to reconstruction, the images are enhanced by removing a background produced by a 5 frame median. For particle counting, the 3D volume is projected to a 2D plane using a maximum intensity projection. The result is then binarized using a threshold equal to 25\% of the maximum intensity. Morphological closing is then used to reduce segmentation noise followed by connected component labelling. The processing steps are summarized in Figure \ref{fig:holo_processing_steps} and a MATLAB implementation of the processing code is available at github.com/HongFFIL/rihvr-matlab.

The compact size of the DIHM sensor enables it to be integrated into a  modified Aquapod to explore a sampling environment while the low cost could allow multiple sensors to be utilized in tandem.
%A sample hologram taken at a depth of 1 m in Madison Lake, MN is shown in Figure \ref{fig:sample_hologram}.

%\begin{figure}[!t]
%\includegraphics[width=6cm]{sample_hologram_scale}
%\includegraphics[width=6cm]{underwater_sample}
%\centering
%\caption{Sample underwater hologram captured using the proposed DIHM sensor}
%\label{fig:sample_hologram}
%\end{figure}

\section{Aquapod} \label{aquapod}

\begin{figure}[!ht]
\centering
\includegraphics[width=1\columnwidth]{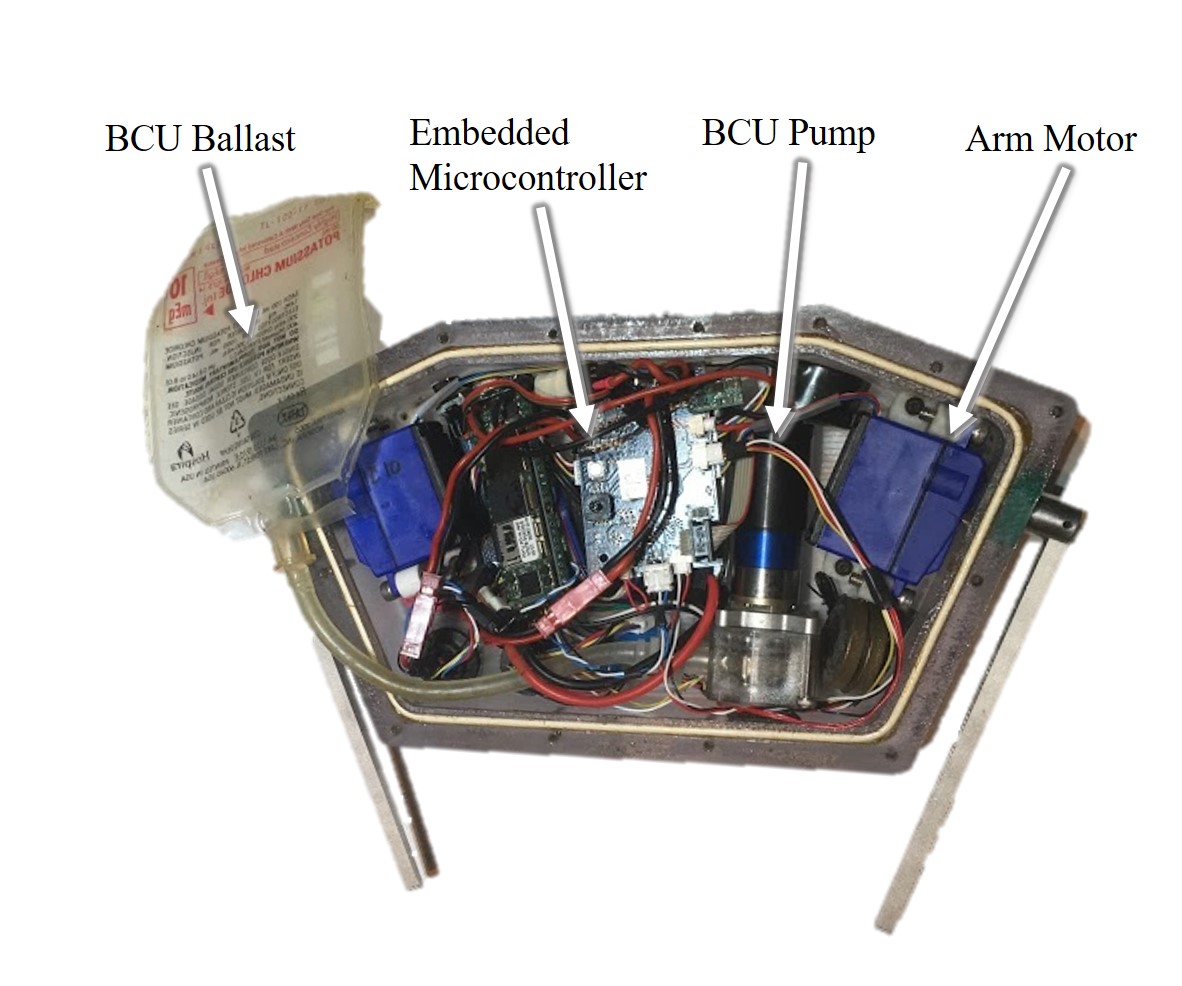}      %photo from diving well day
\caption{Aquapod and its buoyancy control unit (BCU).}
\label{fig:aquapod-callout}
\end{figure}

An amphibious robotic platform from the Center for Distributed Robotics, the Aquapod \cite{Carlson2011b}, employs tumbling locomotion to overcome obstacles that are large when compared to its size. The Aquapod is designed to withstand hydrostatic pressures found at 10 meters of depth. This capability allows it to traverse land or water equally as well as other environments that are difficult for other robots such as snow, sand, or muddy areas. 
These robotic platforms are expected and designed to work in a myriad of environments and overcome unforeseen obstacles, leveraging their submersibility, high trafficability \cite{Hemes2011}, and extendable nature.

An important feature of the robot, the buoyancy control unit (BCU) employs a peristaltic pump to fill a small bladder contained within the robot. This allows for density, and therefore buoyancy manipulation to ascend and descend in a body of water. This \emph{passive} form of locomotion requires less energy to actuate than comparable active methods, such as thrusters, for moving vertically in a column of fluid.  A combination of internal and external pressure sensors allow measurement of depth as well as the amount of ingested fluid residing in the BCU \cite{Dhull2012}.  

%propeller
As can be seen in Figure~\ref{fig:aquapod_dihm}, a set of oppositely oriented and contra-rotating propellers that are independently powered DC motors have been added to the robot to provide lateral movement in the water. The propulsion system is placed far enough afield of the DIHM sensor, seen in the same figure, to minimize fluid entrainment affecting the measured volume. Furthermore, the Aquapod used in Experiment I features propellers and has been modified such that the sensor sits horizontally in the water. 

Despite a minimal amount of hardware complexity, the Aquapod is an inexpensive platform capable of traversing complex terrains with a high mobility-to-size ratio enabling negotiation of a variety of complex terrains where conventional forms of robotic locomotion would fail.

%\begin{figure*}[!h]
%\begin{figure*}[H]
%\centering
%\includegraphics[width=\textwidth{}]{Figures/PointSource_map}
%\caption{Mapped concentration of bubbles near a point source. Insets are enhanced holograms recorded at the marked positions}. 
%\label{fig:concentration_map_point}
%\end{figure*}

%The passive buoyancy control unit (BCU) of the Aquapod enables the use of the DIHM sensor at predetermined depth values that are programmed a priori from the surface. The buoyancy control unit ingests water to increase the robot's mass while maintaining the same rigid hull, in effect increasing its density. Once the density reaches and passes the \emph{neutral buoyancy point} (NBP) it begins the descension process to the predetermined depth. Once the said depth is reached, the BCU is utilized in reverse to decrease the density of the system and begin ascension towards the surface.

%For the purposes of the series of experiments planned, it was desirable to design adjustability into the algorithms to account for payloads of different buoyancy as well as real-world effects not found in a lab setting such as waves from passing boats, naturally occurring waves, a tether, among other effects. In this light, the buoyancy control unit was designed to calibrate by finding its neutral buoyancy point (NBP) during every dive. Once the NBP was established, a closed-loop control was used to drive the Aquapod to a commanded depth. Figure ~\ref{fig:bcu_response} demonstrates the response of the robot in the vertical column subject to the internal pressure.

%propeller

\section{Experiments} \label{experiments}

Two experiments were designed to demonstrate the capabilities of the Aquapod-based DIHM system. First, a laboratory validation experiment was used to demonstrate the system is able to map a concentration field. Second, a field experiment was performed demonstrating the ability of the combined system to measure \emph{in situ} particle concentrations and image aquatic microorganisms in a real-world scenario.

\subsection{Experiment 1: Laboratory Validation}

%\begin{figure*}[!h]
%\begin{figure*}[H]
%\centering
%\includegraphics[width=1\columnwidth]{Figures/PointSource_map}
%\includegraphics[width=\textwidth{}]{Figures/PointSource_map}
%\caption{Mapped concentration of bubbles near a point source. Insets are enhanced holograms recorded at the marked positions}. 
%\label{fig:concentration_map_point}
%\end{figure*}

\begin{figure*}[!h]
\centering
\includegraphics[width=\textwidth{}]{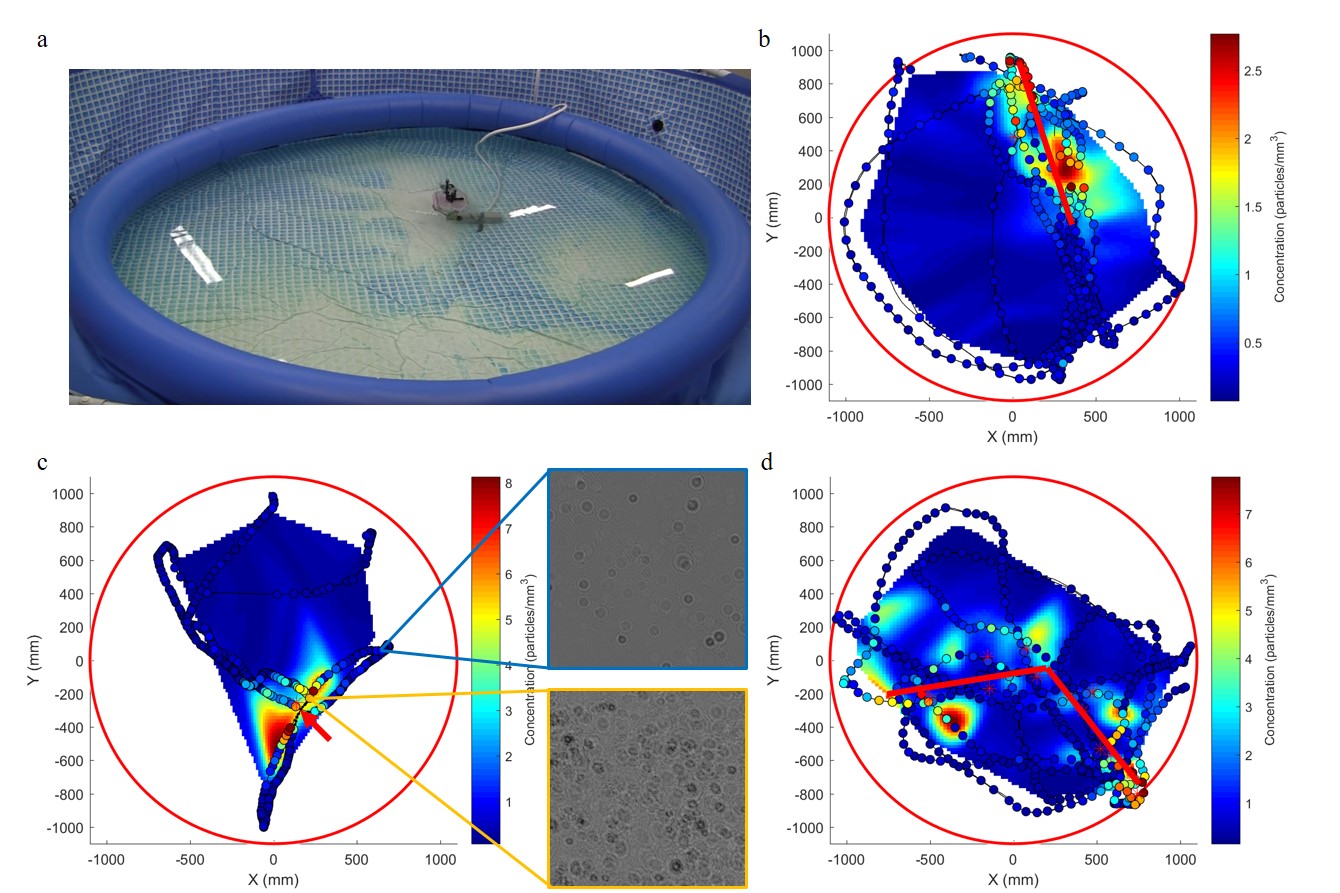}
\caption{(a) The Aquapod with DIHM exploring the point source. (b) Mapped concentration of a linear arrangement of sources. Red line indicates expected source location. (c) Mapped concentration of bubbles near a point source. Red arrow indicates expected source location. Insets are enhanced holograms recorded at the marked positions. (d) Mapped concentration of sources arranged in the bent shape indicated by the red line.}. 
\label{fig:concentration_maps}
\end{figure*}

%\afterpage{\clearpage}

Applications of the autonomous DIHM measurement may include concentration mapping to identify the source of an oil leak or the extent of an HAB. To demonstrate the ability of the robotic DIHM system to map spatial variations in particle concentration, we measured the distribution of bubbles from several source arrangements in a pool. The test volume was a 3 m diameter, 3800 L pool with aerated water sources producing microbubbles to serve as the measured particles (Figure \ref{fig:concentration_maps}(a)). Because the flow rate from the source was constant, the concentration of bubbles is assumed to be steady for the duration of the experiment. The Aquapod was manually driven throughout the pool in a manner which indicated some imprecise knowledge of the source location. The location of the Aquapod and DIHM was tracked using a Vicon Vantage Motion Capture system capable of recording the full attitude of a rigid-object to within 0.1 mm at 100 Hz. Markers for the Vicon can be seen in Figure ~\ref{fig:aquapod_dihm}. Three source configurations were measured. The first was a point source near the center of the pool. The second was a series of five point sources arranged in a line. Finally, six points were arranged in a bent elbow shape.

The 2D path of Aquapod on the water's surface is shown for each configuration in Figure \ref{fig:concentration_maps} along with the mapped concentration. The colored circles identify the locations where the holographic images were recorded and are colored according to the measured particle concentration. A 2D map of the concentration is estimated using a cubic interpolation from the measured points and spatially filtered with a Gaussian filter with a standard deviation of 2. Also shown are several of the recorded holographic images. These images showcase that differences in particulate concentration can be clearly identified even from the unprocessed images. They also exemplify the range of spatial scales which can be measured only using a mobile DIHM system as the holograms show local spatial variations on the sub-millimeter scale while the macro variations are mapped on the meter scale.

\subsection{Experiment 2: \emph{In situ} Measurement}

In order to demonstrate the utility of the combined DIHM-Aquapod system in a real-world scenario, we conducted an \emph{in situ} measurement of particle concentrations in a lake. South Center Lake in Chisago, Minnesota has been identified by the Minnesota Department of Natural Resources as a target of long-term monitoring for trends in biological, chemical, and physical features as a result of human development, climate change, and natural weather patterns. An aquatic monitoring station operated by the Saint Anthony Falls Laboratory was anchored in the lake to record the formation of algal blooms for a full growing season \cite{Wilkinson2020}. The monitoring station utilizes a data sonde (OTT Hydromet, Hydrolab DS5X) which measures water temperature, solar radiation, phycocyanin (a pigment found in some HAB-forming algae), and other properties at 0.5 m depth intervals. Measurements reported by the Sonde are 2 minute averages, recorded every 2 hours. Since the monitoring station is fixed, it is unable to evaluate lateral heterogeneity in the lake. An autonomous robotic system capable of approximating some of the measurements of this station would be a valuable addition to these field experiments.

The Aquapod with DIHM was deployed near the monitoring station to profile the particulate concentration. With the DIHM recording holograms, the Aquapod was commanded to dive to a depth of 5.5 m, remain at that depth, and return to the surface.
%The Aquapod and DIHM clocks are synchronized to enable matching of the recorded holograms to the depth at which they were captured.

%\begin{figure}[!ht]
%\centering
%\includegraphics[width=1\columnwidth]{Figures/aquapod_dihm}
%\caption{DIHM mounted on the Aquapod during deployment.}
%\label{fig:aquapod_dihm}
%\end{figure}

%Due to the complex and non-uniform nature of the particles imaged in the \emph{in situ} environment, a more advanced holographic reconstruction algorithm was necessary. The algorithm is based on the work of Endo et al. \cite{Endo2016} using an iterative regularization algorithm to reduce the noise in the reconstructed image. A sparsity-enforcing regularizer was used and the 2D projection was segmented using Otsu's method. The high computational cost of this method make it impractical in cases with a high number of holograms, leading to its limited use in this case.

%Three depth profiles were measured under the same conditions to ensure that an adequate number of holograms were recorded at all depths. Figure \ref{fig:counted_profile} shows the measurements made by the DIHM compared to the average concentration measured by the measurement station over a 7 day period.

% Due to an unforeseen error, the aquapod depth data was not recorded reliably. For that reason, only the timeseries of concentration data is presented

\begin{figure}[!ht]
\centering
\includegraphics[width=1\columnwidth]{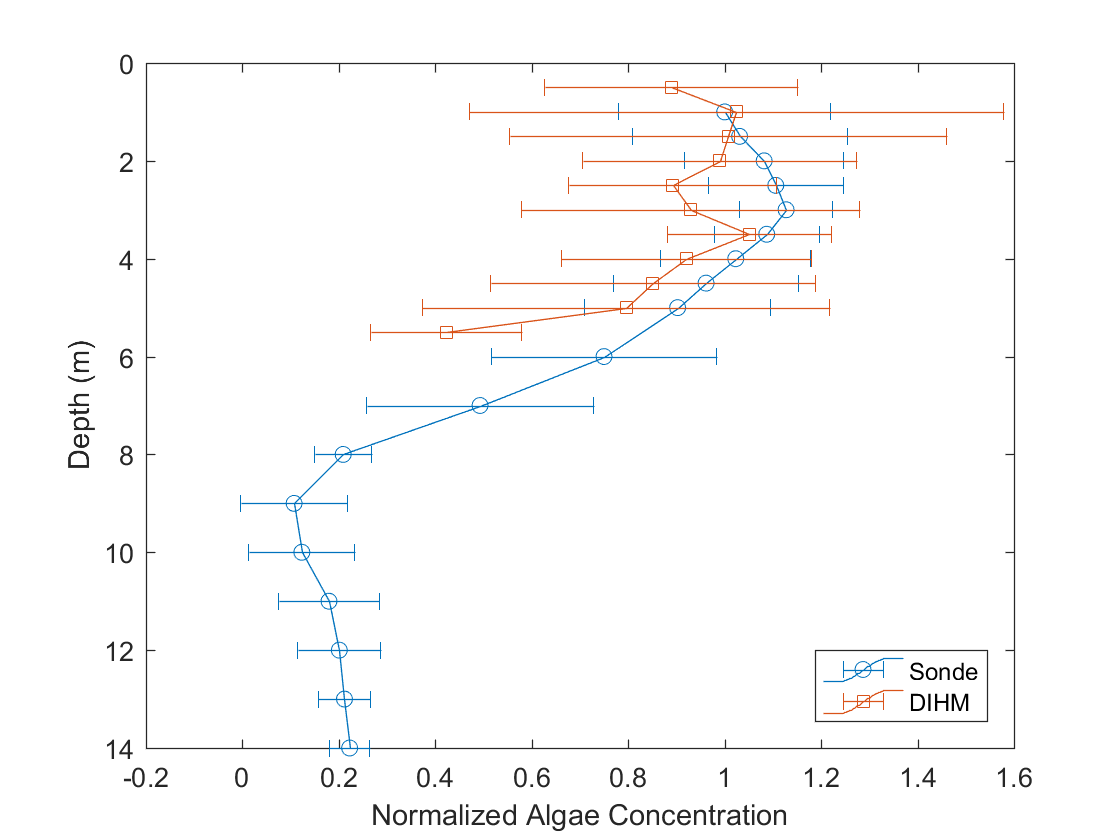}
\caption{Depth profiles of algae concentration from the  sonde and DIHM. Data is normalized by the surface concentration.}
%\caption{Particle concentration from DIHM and depth from Aquapod for a depth profile.}
\label{fig:lake_profile}
\end{figure}

The concentration profiles measured by both the DIHM and the sonde are shown in Figure \ref{fig:lake_profile}. The concentration measurements of the DIHM are binned in 0.5 m increments to match the locations of the sonde measurements. The mean and standard deviation within each of the bins is plotted. The sonde profile is the 7-day average and standard deviation at each depth. All data is normalized using the concentration at the surface and the lake bed (assumed 0 for DIHM). The sonde shows a decrease in algae concentration with depth corresponding to the thermocline -- the region in lakes and oceans characterized by a rapid drop in temperature with depth which also impacts the distribution of microorganisms. The DIHM profile identifies the beginning of this region. The sonde has less standard deviation than the DIHM due to the inherent temporal resolution discrepancy between the methods (sonde is 2 minute average, DIHM is instantaneous). The DIHM is sensitive to small scale spatial and temporal fluctuations in the concentration which the sonde does not record.

%The concentration measured by the DIHM over the duration of the experiment is shown in Figure \ref{fig:lake_profile} along with a selection of holograms. To reduce the impact of occasional poor-quality holograms that result from excessive fluid speed, a 5 second moving average filter is applied. The concentration shows a clear drop as the sensor reaches the maximum depth of the profile. The 7-day average phycocyanin profile produced by the sonde profiler (Figure \ref{fig:phycocyanin}) indicates that the concentration slightly increases from the surface to a maximum at a depth of 3 m. Following that, there is a drop at approximately 6 m corresponding to the thermocline (the region in lakes and oceans characterized by a rapid drop in temperature with depth).  The small peaks at the 125 second and 260 second points may correspond to the 3 m maximum while the sudden drop indicates that the sensor is passing through the thermocline.

%https://v2.overleaf.com/project/5a828b2c4f679c0968aafebd
%\begin{figure}[!ht]
%\centering\includegraphics[width=1\columnwidth]{Figureh%ttps://v2.overleaf.com/project/5a828b2c4f679c0968aafebds/phycocyanin_profile}
%\caption{Phycocyanin depth profile normalized to the surface concentration. Data is averaged over 7 days and error bars are the standard deviation in that time.}
%\label{fig:phycocyanin}
%\end{figure}

\begin{figure}[!ht]
\centering
\includegraphics[width=1\columnwidth]{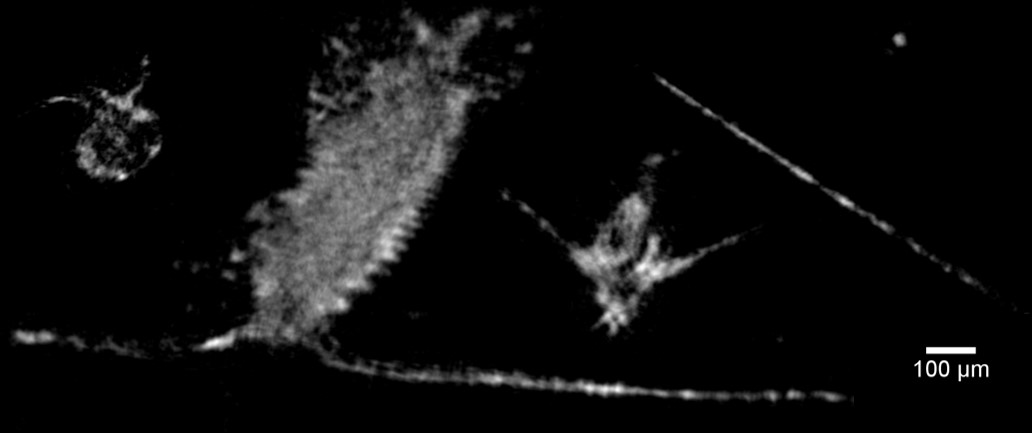}
\caption{Selection of microorganisms imaged during the depth profile.}
\label{fig:morphology}
\end{figure}

A selection of some of the microorganisms seen during the profile are shown in Figure \ref{fig:morphology}. Each object is shown at the manually identified in-focus plane.

%\begin{figure}[!ht]
%\centering
%\includegraphics[width=1\columnwidth]{Figures/counted_particles_profile_norm}
%\caption{Normalized particle concentration profile showing 7-day average from measurement station, 0.5 m average from the DIHM, and number of particles counted in each individual hologram.}
%\label{fig:counted_profile}
%\end{figure}

%\section{Discussion}

%Experiment 1 discussion 

%Figure ~\ref{fig:concentration_map_point} shows the results of experiment 1. Here we can see the path followed by the robot while taking DIHM measurements. 

%Experiment 2 discussion 
%While there is a substantial amount of scatter seen in the DIHM results, the trend of decreasing concentration with depth is clearly visible.

\section{Future Work} \label{future_work}

Presently, the DIHM is capable of recording and storing images which must then be processed externally. The processing throughput is limited by the data transfer and GPU memory bandwidth. The continued development of single board computers, including some with on-board GPU processing, has the potential to enable live processing of holographic images. On-board processing of the recorded holograms will further allow the robot to react accordingly, searching for the peak concentration, and taking water samples there for further analysis.

One could also explore the full capabilities of the DIHM as part of the Aquapod-based design. Automatic classification of aquatic species using a DIHM has been successful both in the laboratory \cite{ElMallahi2013} and in the field using the 4Deep commercial DIHM. However, \emph{in situ} species identification using a low-cost DIHM has not been demonstrated. The 3D nature of holographic reconstructions (see Figure \ref{fig:holo_processing_steps}(b)) was not leveraged for the concentration measurements discussed in this paper. Future studies can examine heterogeneous particle distributions on the micrometer scale as well as examining swimming microorganisms using a faster image capture rate.

Future work will also focus on enhancing the mobility and control of the Aquapod by exploiting the multi-modal capability of the screwdrive to transition from land to the water surface and then into the water column. Such transitions would add further dimensionality to the mission as a wider range of tasks and scenarios could be tackled.  The screwdrive can also be used to achieve more robust depth control as well as faster navigation to desired locations within the body of water with the goal of measuring the heterogeneity of algal concentration in the lakes. 

Enhancing the stability of the robot will also enable higher quality imaging to facilitate measurements that are on par or compete with solutions that are currently employed in the field while allowing researchers more autonomy to carry out experiments outside limits incurred due to human or financial resources.

%\begin{figure*}[!ht]
%\centering
%\includegraphics[width=\textwidth{}]{Figures/future_work.jpg}
%\caption{ (a) Hologram of a dinoflagelate recorded but unidentified during \emph{in situ} measurement. (b) 3D rendering of spherical objects extracted from a hologram. }.
%\label{fig:future_work}
%\end{figure*}

\section{Conclusions}

We have demonstrated a robotic system for \emph{in situ} particle measurements in large water regions at various depths and scales. The DIHM is capable of imaging microscopic particles while the Aquapod is able to traverse larger areas than the state-of-the-art. Integrated in a comprehensive design, they are able to measure at the full range of scales necessary for studies of microorganisms and particulate contaminants. Compared to the existing state-of-the-art, our DIHM is substantially less expensive and smaller -- enabling a broad range of applications including those which utilize multiple DIHMs. We have demonstrated some of the system capabilities with a laboratory experiment mapping the bubble concentration in a pool and an \emph{in situ} measurement of microorganisms in a lake.

%\blindtext

%There is a need for an automated measurement technique to measure the three-dimensional distribution of particle concentrations, such as algal particles, \emph{in situ}. Existing solutions are either limited in scale, or are too expensive for large scale adoption either in terms of hardware cost or human resources. We have proposed a mobile robotic platform using a simple and inexpensive digital in-line holographic microscope  to solve this problem. The DIHM sensor we have developed is able to record holograms underwater with acceptable quality for a fraction of the cost of existing commercial systems. Coupled with an Aquapod, the DIHM has the potential to become an invaluable tool for future water quality research.

% or


%\appendices
%\section{DIHM images}
%\includegraphics[width=6cm]{sample_hologram_scale}
%\centering
%\blindtext

\section*{Acknowledgments}

The authors would like to thank Jiaqi You, Anne Wilkinson, and Miki Hondzo for their knowledge of algae and assistance during deployments. We would also like to thank David Brajkovic for his work developing miniaturized holographic sensors. This material is based upon work supported by the National Science Foundation through grant \#IIS-1427014.

%\ifCLASSOPTIONcaptionsoff
  %\newpage
%\fi



\bibliographystyle{IEEEtran}
\bibliography{references}
%\begin{thebibliography}{1}
%\end{thebibliography}

%\printbibliography
%\bibliographystyle{unsrt}
%\bibliography{references}

\end{document}